\documentclass[letterpaper]{article} 
\usepackage{aaai2027}  
\usepackage[hyphens]{url}  
\usepackage{graphicx} 
\usepackage{natbib}  
\usepackage{caption} 
\usepackage{algorithm}
\usepackage{algorithmic}
\usepackage{amsmath}
\usepackage{amssymb}

\usepackage{newfloat}
\usepackage{listings}
\DeclareCaptionStyle{ruled}{labelfont=normalfont,labelsep=colon,strut=off} 
\floatstyle{ruled}
\newfloat{listing}{tb}{lst}{}
\floatname{listing}{Listing}

\usepackage{booktabs}

\nocopyright 

\title{Scope-WM: Scoped Computation for Efficient Visual World Models}
\author{
Chunzheng Li\textsuperscript{\rm 1},
Zesheng Jia\textsuperscript{\rm 1},
Hongda Zhang\textsuperscript{\rm 2},
Jiaying Tang\textsuperscript{\rm 1},
Yuntian Wang\textsuperscript{\rm 1},
Siao Liu\textsuperscript{\rm 1}\corresponding,
Jin~Wang\textsuperscript{\rm 1}\corresponding
}

\affiliations{
\textsuperscript{\rm 1}Soochow University\\
\textsuperscript{\rm 2}Fudan University\\
}

\begin{document}

\maketitle

\begin{abstract}
Visual world models enable robotic planning by predicting future observations, but dense latent-state propagation and sample-intensive trajectory optimization incur high inference latency and peak memory usage, limiting real-time deployment on resource-constrained platforms.
Existing sparse world-model acceleration methods either rely on unguided token sparsification, which may discard planning-relevant information and
restrict achievable sparsity, or introduce heavy auxiliary modules and cumbersome multi-stage training pipelines.
In this work, we present Scope-WM, an efficient visual world model that scopes computation to prediction-relevant latent regions and promising action sequences.
Scope-WM distills prediction relevance into a lightweight action-conditioned selector and applies full dynamics prediction only to a compact subset of selected tokens.
It updates the remaining tokens using a compact summary of foreground states and their changes, allowing the background to perceive foreground dynamics without costly token-to-token interactions.
During planning, Scope-WM preserves and reuses high-quality action sequences discovered during the initial MPC search, focusing subsequent search under reduced rollout budgets.
The resulting pipeline requires only a one-off selector distillation followed by a single joint training stage for the sparse world model.
On the challenging Push-T task, Scope-WM reduces peak GPU memory usage and planning time to $18.1\%$ and $14.3\%$ of those of dense DINO-WM, respectively, corresponding to a $6.97\times$ planning speedup, while maintaining competitive task performance. Further evaluations across five diverse visual planning tasks demonstrate the general applicability of Scope-WM.
Code is available at \url{https://github.com/ChunZheng2022/Scope-WM}.
\end{abstract}


\section{Introduction}
\label{sec:introduction}

Visual model-based planning provides a general framework for robotic manipulation and control by predicting the consequences of candidate actions before they are executed in the physical environment~\citep{finn2016deep,ebert2018visual,hafner2019planet}.
Rather than learning only a reactive policy, a visual world model predicts future latent states under candidate action sequences and selects actions through model-predictive control (MPC)~\citep{ha2018worldmodels,chua2018pets,hansen2022tdmpc}.
Recent approaches based on self-supervised visual representations, most notably DINO-WM, further enable task-agnostic visual planning by predicting patch-level features extracted from a frozen DINOv2 encoder~\citep{oquab2024dinov2,zhou2025dinowm}.
However, this patch-level formulation is computationally demanding.
At every MPC step, the model repeatedly propagates all visual tokens over long prediction horizons for many action sequences sampled by the cross-entropy method (CEM)~\citep{de2005tutorialcem}.
Consequently, the cost of latent imagination grows with both the number of visual tokens and the rollout population, resulting in high planning latency and peak memory usage that hinder real-time deployment.

\begin{figure}[!t]
    \centering
    \includegraphics[width=0.98\columnwidth]{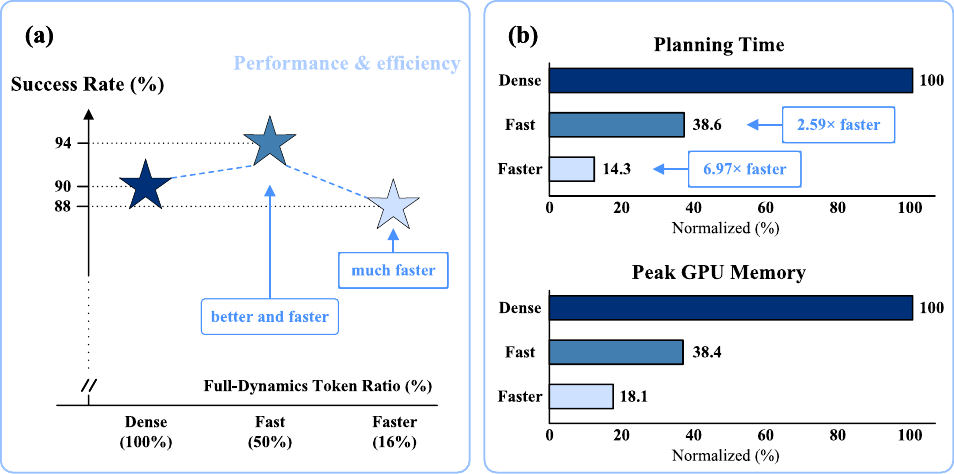}
    \caption{
    Performance--efficiency trade-off of Scope-WM on Push-T.
    (a) Different token budgets yield different performance--efficiency operating points under varying sparsity.
    (b) Scope-WM substantially reduces planning time and peak GPU memory over dense DINO-WM.
    }
    \label{fig:scope_tradeoff}
\end{figure}

A natural approach to reducing this cost is to make latent imagination sparse by applying full dynamics prediction to only a compact subset of visual tokens.
Existing sparse world-model acceleration methods either rely on unguided
token sparsification, which may discard planning-relevant information and
limit the sparsity that can be applied safely, or introduce specialized
auxiliary modules and cumbersome multi-stage training pipelines~\citep{chun2026sparseimagination,yin2026ddpwm}.
More fundamentally, efficient sparse imagination involves more than deciding how many tokens to retain.
It must determine which tokens warrant accurate dynamics prediction, how the remaining tokens should evolve without eliminating the computational savings, and how cheaper latent rollouts can be translated into more efficient trajectory optimization.
These decisions are tightly coupled: inaccurate token allocation can degrade future-state prediction, expensive background modeling can offset the savings from sparsity, and sample-intensive planning can remain a dominant bottleneck even when individual rollouts become cheaper.

In this work, we present Scope-WM, an efficient visual world model that scopes computation to prediction-relevant latent regions and promising action sequences.
Scope-WM allocates computation at three complementary levels.
At the token level, it identifies a compact subset of tokens that require full dynamics prediction.
At the dynamics level, it updates the remaining tokens using lightweight foreground-conditioned computation rather than dense token-to-token interactions.
At the planning level, it concentrates subsequent trajectory search around high-quality action sequences discovered during the initial MPC search.
As illustrated in Figure~\ref{fig:scope_tradeoff}, this scoped design provides flexible operating points across different compute budgets: a moderate sparse configuration can improve efficiency while preserving or even improving task performance, whereas a more aggressive configuration prioritizes latency and memory reduction while retaining competitive control performance.

Concretely, these three levels are realized by Distilled Relevance Selection
(DRS), Foreground-Delta Background Update (FDBU), and Elite-Bank CEM
(EB-CEM), respectively.
The resulting pipeline requires only a one-off selector distillation followed
by a single joint training stage for the sparse dynamics predictor and FDBU.
Evaluations demonstrate that Scope-WM substantially improves planning efficiency while maintaining competitive task performance.

Our main contributions are summarized as follows:
\begin{itemize}
    \item We formulate efficient visual world-model planning as a three-level computation-allocation problem spanning prediction-relevant token selection, foreground-aware background updating, and trajectory-search budgeting.

    \item We propose DRS, a lightweight action-conditioned selector that distills prediction relevance from a full-token teacher to focus full dynamics prediction on a compact token subset. DRS requires only one-off distillation, followed by joint training of the sparse predictor and FDBU.

    \item We propose FDBU, which summarizes foreground states and deltas to perform efficient token-wise background updates without dense foreground--background interactions.

    \item We identify the path-dependent behavior of low-budget CEM planning and propose EB-CEM, which front-loads the sampling budget and persistently reuses high-quality initial action sequences across MPC steps. Together, these components reduce both per-trajectory imagination cost and subsequent trajectory-search cost while preserving competitive control performance.
\end{itemize}

\section{Related Work}
\label{sec:related_work}

\paragraph{Visual world models and pretrained representations.}
Visual model-based control initially predicted future observations in pixel space and optimized action sequences through MPC~\citep{finn2016deep,ebert2018visual}.
Subsequent methods moved prediction into compact latent spaces: World Models, PlaNet, and Dreamer learn recurrent latent dynamics for control through imagined trajectories, while PETS and TD-MPC combine learned dynamics with sampling-based or temporally abstract planning objectives~\citep{ha2018worldmodels,hafner2019planet,hafner2019dreamer,hafner2023dreamerv3,chua2018pets,hansen2022tdmpc,hansen2024tdmpc2}.
Transformer- and diffusion-based approaches, including IRIS, Transformer-based World Models, STORM, Masked World Models, and DIAMOND, further improve latent sequence modeling and visual prediction~\citep{micheli2023iris,robine2023twm,zhang2023storm,seo2023maskedworldmodels,alonso2024diamond}.
In parallel, pretrained representations such as R3M, MVP, and VIP provide reusable features or goal-conditioned reward signals for control, while DINO, DINOv2, I-JEPA, and V-JEPA learn general spatial or temporal visual representations through self-supervision~\citep{nair2022r3m,xiao2022mvp,radosavovic2023realmvp,ma2023vip,caron2021dino,oquab2024dinov2,assran2023ijepa,bardes2024vjepa,assran2025vjepa2}.
DINO-WM combines these directions by predicting future patch-level DINOv2 features and planning directly in the resulting feature space~\citep{zhou2025dinowm}.
More recently, LeWorldModel learns compact global latent representations end-to-end from pixels, reducing planning cost by changing the representation and training paradigm itself~\citep{maes2026lewm}.
Scope-WM instead follows the pretrained patch-level formulation of DINO-WM and improves its efficiency by selectively allocating computation within spatially structured latent rollouts and trajectory search.

\paragraph{Sparse visual computation and imagination.}
Efficient Vision Transformers exploit token redundancy through learned pruning, adaptive computation, token reorganization, or token merging to reduce inference cost~\citep{rao2021dynamicvit,xu2022evovit,yao2026vpruner,liang2022evit,yin2022avit,bolya2023tome}.
These methods are primarily designed for static recognition, whereas visual world-model planning requires state- and action-dependent token selection whose errors may accumulate over multi-step rollouts.
Sparse Imagination applies random patch dropping and randomized grouped attention to DINO-style world models, demonstrating that moderate sparsity can preserve planning performance while avoiding the spatial bias of fixed importance scores~\citep{chun2026sparseimagination}.
However, random selection remains independent of the current prediction objective.
DDP-WM instead decomposes latent dynamics into sparse primary dynamics and context-driven background updates, using a dynamic localization network for foreground identification and foreground-conditioned cross-attention for background correction~\citep{yin2026ddpwm}.
Scope-WM shares the goal of separating expensive foreground prediction from inexpensive background maintenance, but adopts lighter mechanisms: DRS distills prediction relevance into a compact token-wise selector, while FDBU replaces pairwise foreground--background interactions with a pooled foreground-state and foreground-delta context.
After selector distillation, the sparse predictor and background updater can therefore be optimized in a unified training stage.

\paragraph{Sampling-based planning.}
CEM is widely used with learned dynamics models because it can optimize action sequences without differentiating through the planning objective, but its cost scales directly with the number of candidate rollouts~\citep{de2005tutorialcem,chua2018pets}.
iCEM improves sample efficiency through temporally correlated noise, warm starts, and elite reuse across optimization iterations and planning steps~\citep{pinneri2020icem}.
EB-CEM is complementary to these techniques and focuses on the allocation of samples across receding-horizon MPC stages.
It assigns a larger population to the initial broad search, stores high-quality early trajectories in a persistent bank, and
samples perturbed variants during later low-budget replanning.
\section{Method}
\label{sec:method}

\begin{figure*}[!t]
    \centering
    \includegraphics[width=0.98\textwidth]{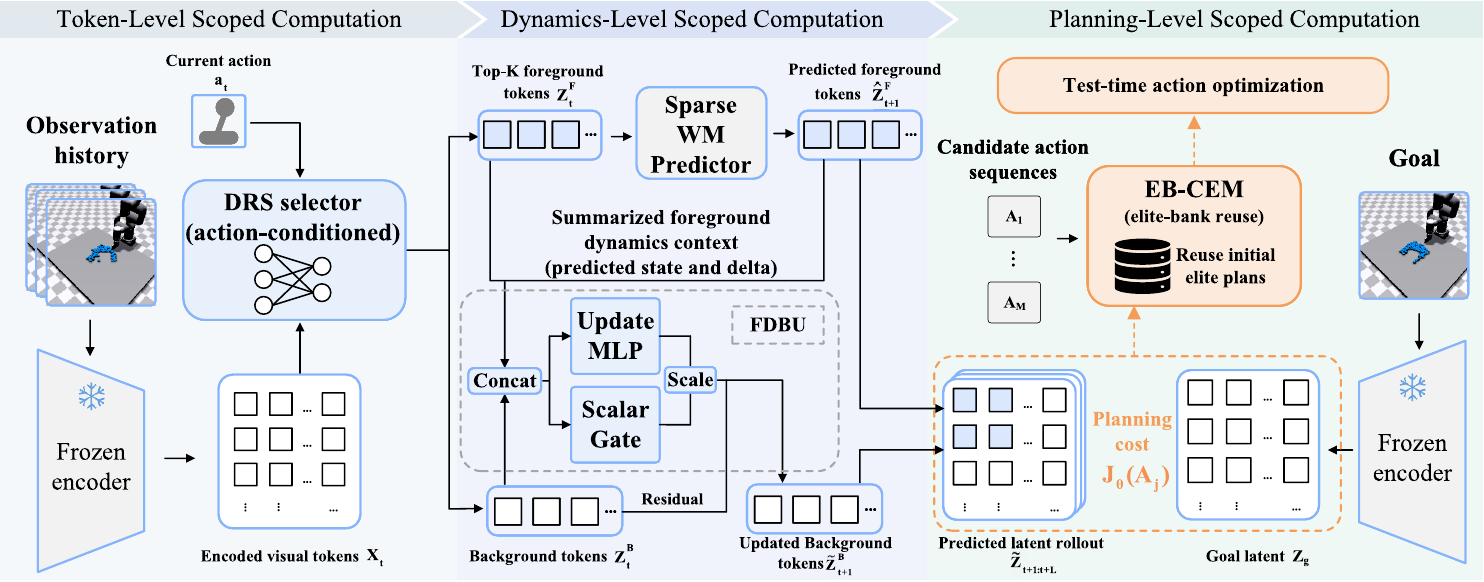}
    \caption{
    Overview of Scope-WM.
    Scope-WM scopes computation at the token, dynamics, and planning levels.
    At the token level, the action-conditioned DRS selector identifies a compact set of prediction-relevant foreground tokens from the encoded visual state.
    At the dynamics level, the sparse world-model predictor applies full dynamics prediction to the selected tokens, while FDBU uses a summarized foreground-dynamics context to update the remaining background tokens through a lightweight gated residual update, producing the complete latent rollout.
    At the planning level, EB-CEM performs test-time action optimization by retaining and reusing high-quality elite action sequences across MPC steps.
    Candidate action sequences are evaluated by a planning objective that compares the predicted latent rollout with the goal latent encoded by the frozen visual encoder.
    }
    \label{fig:scope_overview}
\end{figure*}

We build Scope-WM on top of DINO-WM~\citep{zhou2025dinowm} and describe
its token-, dynamics-, and planning-level components in turn, as illustrated
in Figure~\ref{fig:scope_overview}.

Given an observation \(o_t\), the frozen DINOv2 encoder extracts a set of \(N\) visual patch tokens
\begin{equation}
    X_t = \{x_{t,i}\}_{i=1}^{N}, \qquad x_{t,i}\in\mathbb{R}^{d_x}.
\end{equation}
The visual tokens are combined with proprioceptive state \(p_t\) and action embedding \(e(a_t)\) to construct world-model tokens
\begin{equation}
    z_{t,i} = \mathrm{Fuse}(x_{t,i}, p_t, e(a_t)) \in \mathbb{R}^{d_z}.
\end{equation}
Given a history length \(H\), the dense world model predicts future latent tokens conditioned on the observed history as
\begin{equation}
    \hat{Z}_{2:H+1} = f_\theta(Z_{1:H}),
\end{equation}
where \(Z_t=\{z_{t,i}\}_{i=1}^{N}\). 
The dense predictor operates on all \(HN\) tokens, which is expensive during MPC because CEM requires evaluating many candidate action sequences per step.
Our method reduces this cost through three components: Distilled Relevance Selection, Foreground-Delta Background Update, and Elite-Bank CEM.

\subsection{Distilled Relevance Selection}
\label{sec:drs}

A central challenge in sparse latent imagination is deciding which tokens should be kept.
Prior work has observed that static attention-based token selection is often insufficient for visual planning~\citep{chun2026sparseimagination}.
This is because prediction-relevant regions are not fixed across successive
planning steps: the relevant tokens depend on the current state, the planned
action, and the prediction objective.
A purely visual-difference-based selector can also be problematic.
Although frame-level changes can highlight moving regions, they may ignore
tokens that are dynamically important but visually stable, such as contacts,
obstacles, or structures that constrain future motion.
Therefore, we use a dynamic, prediction-aware selection mechanism.

\paragraph{Teacher relevance target.}
Starting from a trained dense DINO-WM teacher, we compute its latent prediction loss:
\begin{equation}
    \mathcal{L}_{\mathrm{T}}
    =
    \frac{1}{|\Omega|}
    \sum_{(t,i)\in\Omega}
    \left\|
        \hat{z}^{\mathrm{T}}_{t,i}
        -
        z^{\star}_{t,i}
    \right\|_2^2 ,
    \label{eq:teacher_loss}
\end{equation}
where \(\hat{z}^{\mathrm{T}}_{t,i}\) is the teacher prediction, \(z^{\star}_{t,i}\) is the target latent token, and \(\Omega\) denotes the supervised token set. 
We then estimate the contribution of each visual token to the teacher prediction loss using a gradient-times-input score:
\begin{equation}
    \tilde{r}_{t,i}
    =
    \left\|
        x_{t,i}
        \odot
        \frac{\partial \mathcal{L}_{\mathrm{T}}}{\partial x_{t,i}}
    \right\|_2 .
    \label{eq:grad_input}
\end{equation}
The score is normalized within each frame:
\begin{equation}
    q_{t,i}
    =
    \frac{
        \tilde{r}_{t,i}+\epsilon
    }{
        \sum_{j=1}^{N}(\tilde{r}_{t,j}+\epsilon)
    } .
    \label{eq:teacher_relevance_distribution}
\end{equation}
The resulting distribution \(q_t\) serves as a soft relevance target.
Unlike static attention masks or visual-change heuristics, this target directly reflects which tokens affect the world model's prediction error under the current objective.

\paragraph{Lightweight relevance head.}
We train a compact relevance head \(g_\phi\) to predict token relevance scores.
For each token, the visual feature and control context are jointly projected into
a shared hidden space:
\begin{equation}
h_{t,i}
=
\phi_x(x_{t,i})
+
\phi_p(p_t)
+
\phi_a(e(a_t)),
\qquad
s_{t,i}=g_\phi(h_{t,i}),
\end{equation}
where \(\phi_x\), \(\phi_p\), and \(\phi_a\) are lightweight linear
projections, and \(g_\phi\) is a small MLP producing the relevance logit.
The predicted relevance distribution across tokens is obtained by a temperature-scaled softmax:
\begin{equation}
    \pi_{t,i}
    =
    \frac{
        \exp(s_{t,i}/T)
    }{
        \sum_{j=1}^{N}\exp(s_{t,j}/T)
    } ,
    \label{eq:student_relevance_distribution}
\end{equation}
where \(T\) is the distillation temperature.
The DRS head is trained by minimizing the KL divergence from the corresponding teacher relevance target:
\begin{equation}
    \mathcal{L}_{\mathrm{DRS}}
    =
    D_{\mathrm{KL}}(q_t \,\|\, \pi_t)
    =
    \sum_{i=1}^{N}
    q_{t,i}
    \log
    \frac{q_{t,i}}{\pi_{t,i}} .
    \label{eq:drs_loss}
\end{equation}
After distillation, the selector is fixed during sparse world-model training.
At inference time, DRS selects the top-\(K\) tokens according to their predicted relevance:
\begin{equation}
    m_{t,i}
    =
    \mathbb{I}
    \left[
        s_{t,i}
        \in
        \mathrm{TopK}
        \left(
            \{s_{t,j}\}_{j=1}^{N}, K
        \right)
    \right],
    \label{eq:topk_mask}
\end{equation}
where \(m_{t,i}\in\{0,1\}\) is the foreground mask.
The resulting foreground and background token sets are
\begin{equation}
    \mathcal{F}_t=\{i\mid m_{t,i}=1\},
    \qquad
    \mathcal{B}_t=\{i\mid m_{t,i}=0\}.
\end{equation}

The selector is deliberately lightweight. Since frozen DINOv2
patch tokens are already globally contextualized through self-attention,
DRS can exploit image-wide visual context without an additional
ViT-style mask generator. It therefore uses only a small token-wise
MLP conditioned on state and action, introducing negligible rollout
overhead while producing state-, action-, and prediction-aware masks.

\subsection{Foreground-Delta Background Update}
\label{sec:fdbu}

After selecting foreground tokens, a natural approach is to run the world model only on these tokens and keep the background tokens unchanged. 
However, this may cause latent inconsistency: background tokens may still need small corrections due to object motion, contact, occlusion, or camera-frame changes. 
One possible solution is to use an attention-based background update module, but such a design increases computational cost and often requires staged training. 
We instead introduce a simpler module, Foreground-Delta Background Update,
which is lightweight and can be trained jointly with the sparse world model.

Let \(Z^F_t=\{z_{t,i}\mid i\in\mathcal{F}_t\}\) denote the selected
foreground tokens. The primary sparse dynamics predictor is applied only
to these tokens during latent rollout:
\begin{equation}
    \hat{Z}^{F}_{2:H+1}
    =
    f_\theta(Z^{F}_{1:H}).
    \label{eq:sparse_foreground_prediction}
\end{equation}
For a Transformer predictor, this reduces the dominant attention cost from \(O((HN)^2)\) to \(O((HK)^2)\), where \(K\ll N\).

FDBU updates the remaining background tokens using the predicted foreground dynamics.
For each predicted step, we compute the foreground delta:
\begin{equation}
    \Delta^F_{t+1,i}
    =
    \hat z^F_{t+1,i}-z^F_{t,i},
    \qquad i\in\mathcal{F}_t.
    \label{eq:foreground_delta}
\end{equation}
We summarize the predicted foreground state and foreground
motion separately by mean pooling:
\begin{equation}
    \bar{z}^{F}_{t+1}
    =
    \frac{1}{K}
    \sum_{i\in\mathcal{F}_t}
    \hat{z}^{F}_{t+1,i},
    \qquad
    \bar{\Delta}^{F}_{t+1}
    =
    \frac{1}{K}
    \sum_{i\in\mathcal{F}_t}
    \Delta^{F}_{t+1,i}.
    \label{eq:foreground_summary}
\end{equation}
The two summaries are fused into a foreground dynamics context to guide subsequent background updates:
\begin{equation}
    c_{t+1}
    =
    \psi_c
    \left(
        [\bar{z}^{F}_{t+1}, \bar{\Delta}^{F}_{t+1}]
    \right),
    \label{eq:fdbu_context}
\end{equation}
where \(\psi_c\) is a small projection network.

For each full-grid token \(z_{t,i}\), FDBU independently predicts a lightweight next-step residual update and scalar gate using the foreground dynamics context:
\begin{equation}
    u_{t+1,i}
    =
    \psi_u([z_{t,i}, c_{t+1}]),
    \qquad
    \gamma_{t+1,i}
    =
    \sigma(\psi_g([z_{t,i}, c_{t+1}])) ,
    \label{eq:fdbu_residual_gate}
\end{equation}
where \(\psi_u\) and \(\psi_g\) are lightweight MLPs and \(\sigma(\cdot)\) is the sigmoid function.
Only background tokens receive this update:
\begin{equation}
    \tilde{z}^{B}_{t+1,i}
    =
    z_{t,i}
    +
    \alpha
    (1-m_{t,i})
    \gamma_{t+1,i}
    u_{t+1,i},
    \label{eq:fdbu_background_update}
\end{equation}
where \(\alpha\) controls the residual update scale.
The final full-grid prediction is formed by scattering foreground predictions into the full grid and applying FDBU to the background:
\begin{equation}
    \tilde{z}_{t+1,i}
    =
    m_{t,i}\hat{z}^{F}_{t+1,i}
    +
    (1-m_{t,i})\tilde{z}^{B}_{t+1,i}.
    \label{eq:fdbu_scatter}
\end{equation}

The sparse world model is trained with a full-grid latent prediction loss over all latent positions:
\begin{equation}
\mathcal{L}_{\mathrm{WM}}
=
\operatorname{MSE}\!\left(\widetilde{Z}, Z^\star\right),
\end{equation}
which jointly supervises foreground prediction and FDBU background updates during sparse-model training.
FDBU is integrated into the world-model training pipeline and does not
require a separate training stage.

\subsection{Elite-Bank CEM}
\label{sec:ebcem}

CEM-based MPC is sensitive to the number of sampled action sequences. 
Empirically, we observe that increasing the number of samples can significantly improve success rate in the low-sample regime, but the gain quickly saturates when the sample count becomes large. 
Moreover, later MPC steps often provide limited recovery ability: once early actions move the agent or object into an unfavorable state, subsequent replanning may not be able to recover. 
This suggests that, in this regime, improving the quality of early candidate action sequences is more important than simply increasing the number of samples at every step.

Motivated by this observation, we introduce Elite-Bank CEM, a simple proposal-reuse mechanism for MPC.
It leaves the planning objective and world model unchanged.
Instead, it improves later CEM proposals by reusing high-quality action sequences found during the initial MPC search.

Let \(A_j\in\mathbb{R}^{L\times d_a}\) be a candidate action sequence of planning horizon \(L\), where \(j=1,\ldots,M\).
At CEM iteration \(\ell\), standard CEM samples candidates from a diagonal Gaussian:
\begin{equation}
    A_j^{(\ell)}
    \sim
    \mathcal{N}
    \left(
        \mu^{(\ell)},
        \mathrm{diag}\left((\sigma^{(\ell)})^2\right)
    \right).
    \label{eq:cem_sampling}
\end{equation}
Each candidate is evaluated by the world-model rollout cost:
\begin{equation}
    C_j^{(\ell)}
    =
    J_{\theta}(A_j^{(\ell)}),
    \label{eq:cem_cost}
\end{equation}
where lower cost indicates a better action sequence. 
The top \(K_{\mathrm{elite}}\) candidates form the elite set:
\begin{equation}
    \mathcal{E}^{(\ell)}
    =
    \mathrm{TopK}_{\mathrm{min}}
    \left(
        \{A_j^{(\ell)}\}_{j=1}^{M},
        \{C_j^{(\ell)}\}_{j=1}^{M},
        K_{\mathrm{elite}}
    \right).
    \label{eq:cem_elite_set}
\end{equation}
CEM then updates its Gaussian parameters:
\begin{equation}
    \mu^{(\ell+1)}
    =
    \frac{1}{K_{\mathrm{elite}}}
    \sum_{A\in\mathcal{E}^{(\ell)}} A,
    \label{eq:cem_mean_update}
\end{equation}
\begin{equation}
    \sigma^{(\ell+1)}
    =
    \mathrm{Std}
    \left(
        \mathcal{E}^{(\ell)}
    \right).
    \label{eq:cem_std_update}
\end{equation}

EB-CEM front-loads search by using a larger candidate population at the
initial MPC step and a reduced population thereafter to emphasize early-stage exploration. Let \(M_r\) denote
the candidate budget at MPC step \(r\):
\begin{equation}
M_r =
\begin{cases}
M_{\mathrm{front}}, & r=0,\\
M_{\mathrm{later}}, & r>0,
\end{cases}
\qquad
M_{\mathrm{front}}>M_{\mathrm{later}}.
\end{equation}
We use \(M_{\mathrm{front}}=300\) and \(M_{\mathrm{later}}=100\) by default.

After the initial MPC step, we store the best \(M_b\) candidates from its
final CEM iteration in a persistent elite bank:
\begin{equation}
\mathcal{B}
=
\operatorname{TopM}_{\min}
\left(
\{A_j\}_{j=1}^{M_{\mathrm{front}}},
\{J_\theta(A_j)\}_{j=1}^{M_{\mathrm{front}}},
M_b
\right).
\end{equation}

At each CEM iteration of subsequent MPC steps, a fraction \(\eta\) of the
reduced population is sampled locally around the banked elites to refine promising solutions:
\begin{equation}
M_{\mathrm{local}}=\lfloor \eta M_{\mathrm{later}}\rfloor,\qquad
M_{\mathrm{global}}=M_{\mathrm{later}}-M_{\mathrm{local}}.
\end{equation}
For a banked elite sequence \(B\in\mathcal{B}\), a local candidate is
generated by Gaussian perturbation:
\begin{equation}
A_{\mathrm{local}}
=
B+\lambda\sigma^{(\ell)}\odot\xi,
\qquad
\xi\sim\mathcal{N}(0,I),
\end{equation}
where \(\lambda\) controls the perturbation scale. The remaining candidates
are instead sampled globally:
\begin{equation}
A_{\mathrm{global}}
\sim
\mathcal{N}\!\left(
\mu^{(\ell)},
\operatorname{diag}\!\left((\sigma^{(\ell)})^2\right)
\right).
\end{equation}
The final candidate set combines the local and global samples to balance local refinement and global exploration:
\begin{equation}
\mathcal{A}
=
\mathcal{A}_{\mathrm{local}}
\cup
\mathcal{A}_{\mathrm{global}}.
\end{equation}
This mechanism encourages the planner to preserve and refine promising early plans, while still keeping global exploration through CEM samples. 
Therefore, EB-CEM is complementary to the sparse world model: DRS and FDBU reduce the cost of each rollout, while EB-CEM improves search quality by reallocating the rollout budget across MPC steps.

\section{Experiments}
\label{sec:experiments}

\subsection{Setup}
\label{sec:experimental_setup}

\paragraph{Environments and datasets.}
We evaluate Scope-WM on five visual planning tasks:
PointMaze~\citep{fu2021d4rl}, Wall,
Push-T~\citep{chi2023diffusionpolicy},
Rope~\citep{zhang2024adaptigraph}, and Granular,
following the task definitions and offline datasets of
DINO-WM~\citep{zhou2025dinowm}.
These tasks cover navigation, rigid-body manipulation, and
deformable or multi-body dynamics across diverse physical interaction regimes.

\paragraph{Evaluation metrics.}
Following DINO-WM, all tasks use CEM-based MPC over latent world-model
rollouts.
We report success rate (SR, $\uparrow$) for PointMaze, Wall, and Push-T,
and Chamfer Distance (CD, $\downarrow$) for Rope and Granular.
For efficiency, we report three complementary metrics: transition-model FLOPs, planning time, and peak
GPU memory usage. Planning time is the cumulative CEM sub-planner time
across MPC steps, while peak memory is the maximum CUDA memory allocated
during each evaluation run.

\paragraph{Baseline.}
We use dense DINO-WM~\citep{zhou2025dinowm} as the baseline for direct comparison.
Both methods share the encoder, data, evaluation pairs, planning objective,
horizon, and action constraints unless explicitly noted.

\paragraph{Implementation details.}
PointMaze, Wall, and Push-T are trained using eight NVIDIA V100 32G GPUs,
with planning evaluated on a single NVIDIA V100 GPU.
Rope and Granular are trained and evaluated on a single NVIDIA RTX
2080 Ti GPU due to PyFlex software compatibility.
Additional implementation details for training and planning are included in the supplementary material.

\begin{table*}[t]
    \centering
    {
    \small
    \setlength{\tabcolsep}{4.0pt}
    \begin{tabular}{lccccccc}
        \toprule
        Method
        & Full-dyn. tokens
        & SR (\(\uparrow\))
        & FLOPs (G, \(\downarrow\))
        & Plan time (s, \(\downarrow\))
        & Speedup (\(\uparrow\))
        & Peak Mem. (MB, \(\downarrow\))
        & Mem. Ratio (\(\downarrow\)) \\
        \midrule
        DINO-WM (Dense)
        & 196
        & 90\%
        & 31.89
        & 240.96
        & \(1.00\times\)
        & 18436
        & 100.0\% \\

        Scope-WM (Fast)
        & 98
        & \textbf{94\%}
        & 15.04
        & 93.06
        & \(2.59\times\)
        & 7088
        & 38.4\% \\

        Scope-WM (Faster)
        & 32
        & 88\%
        & \textbf{5.27}
        & \textbf{34.56}
        & \(\mathbf{6.97}\times\)
        & \textbf{3334}
        & \textbf{18.1\%} \\
        \bottomrule
    \end{tabular}
    }
    \caption{
        End-to-end performance--efficiency trade-off on Push-T.
        Fast and Faster apply full dynamics prediction to \(98\) and
        \(32\) of \(196\) visual tokens.
        FLOPs are reported per prediction sample, with memory
        ratios normalized to DINO-WM.
    }
    \label{tab:end_to_end_tradeoff}
\end{table*}

\begin{table*}[t]
    \centering
    {
    \small
    \setlength{\tabcolsep}{7pt}
    \begin{tabular}{lccccc}
        \toprule
        Method
        & PointMaze (SR \(\uparrow\))
        & Push-T (SR \(\uparrow\))
        & Wall (SR \(\uparrow\))
        & Rope (CD \(\downarrow\))
        & Granular (CD \(\downarrow\)) \\
        \midrule
        IRIS
        & 74\%
        & 32\%
        & 4\%
        & 1.11
        & 0.37 \\

        DreamerV3
        & \textbf{100\%}
        & 30\%
        & \textbf{100\%}
        & 2.49
        & 1.05 \\

        Sparse Imagination
        & \textbf{100\%}
        & 78.3\%
        & 95\%
        & --
        & -- \\

        DINO-WM
        & 98\%
        & 90\%
        & 96\%
        & \textbf{0.41}
        & 0.26 \\
        \midrule

        Scope-WM (Fast, \(K=98\))
        & \textbf{100\%}
        & \textbf{94\%}
        & 92\%
        & 0.87
        & \textbf{0.22} \\

        Scope-WM (Faster, \(K=32\))
        & 98\%
        & 88\%
        & 92\%
        & 0.71
        & 0.25 \\
        \bottomrule
    \end{tabular}
    }
    \caption{
        Cross-task planning performance.
        SR (\(\uparrow\)) is reported for PointMaze, Push-T, and Wall,
        and CD (\(\downarrow\)) for Rope and Granular.
        Fast and Faster use \(98\) and \(32\) full-dynamics tokens.
    }
    \label{tab:planning_performance}
\end{table*}

\subsection{Main Results}
\label{sec:main_results}

We first evaluate the end-to-end performance and efficiency of Scope-WM on Push-T, and then examine its planning performance across all five visual planning tasks.

\paragraph{End-to-end performance and efficiency.}
Table~\ref{tab:end_to_end_tradeoff} compares dense DINO-WM with two
Scope-WM operating points on Push-T.
Fast applies full dynamics prediction to \(98\) of \(196\) visual
tokens, improving SR from \(90\%\) to \(94\%\) while reducing
per-sample FLOPs from \(31.89\) to \(15.04\) G.
Its planning time and peak GPU memory decrease to \(38.6\%\) and
\(38.4\%\) of the dense baseline, respectively, yielding a
\(2.59\times\) planning speedup.
Faster further reduces full dynamics prediction to \(32\) tokens and
requires only \(5.27\) G FLOPs per sample, a \(6.05\times\) reduction
over dense DINO-WM.
It retains \(88\%\) SR while reducing planning time and peak memory to
\(14.3\%\) and \(18.1\%\), respectively, corresponding to a
\(6.97\times\) planning speedup.
These results show that simply adjusting the number of selected
top-ranked tokens allows Scope-WM to either improve both planning
performance and efficiency, or achieve substantially greater acceleration
while retaining competitive task performance.

\paragraph{Planning performance across tasks.}
Table~\ref{tab:planning_performance} compares Scope-WM with prior
methods across five tasks.
Under the aggressive \(K=32\) setting, Scope-WM achieves \(98\%\),
\(88\%\), and \(92\%\) SR on PointMaze, Push-T, and Wall, matching
DINO-WM on PointMaze and remaining within \(2\) and \(4\) percentage
points on Push-T and Wall, respectively.
It achieves a CD of \(0.25\) on Granular, slightly improving over
DINO-WM at \(0.26\).
With the larger \(K=98\) budget, Fast reaches \(100\%\) on PointMaze,
\(94\%\) on Push-T, and \(0.22\) CD on Granular.
Rope remains the most challenging task for the sparse formulation,
where the best Scope-WM result is \(0.71\) CD, compared with \(0.41\)
for dense DINO-WM.
Overall, Scope-WM preserves strong planning performance on most tasks
while substantially reducing latent-rollout computation.

\subsection{Ablation Study}
\label{sec:ablation}

We conduct ablation studies on Push-T to examine the contribution of
the proposed designs and how they interact across model prediction and
trajectory optimization.

\paragraph{Component ablation.}
Table~\ref{tab:component_ablation} first isolates the contribution of
the three components under the aggressive 32-token setting.
Replacing DRS with random token selection reduces the success rate from
\(88\%\) to \(76\%\), showing the importance of allocating full
dynamics prediction to relevant regions.
Removing FDBU and directly carrying background tokens forward reduces
the success rate to \(56\%\), indicating that unselected tokens still
require lightweight adaptation to foreground dynamics.
Replacing EB-CEM with standard CEM decreases the success rate to
\(82\%\).
The complete Scope-WM achieves the highest success rate, showing that
the three designs provide complementary benefits.

\begin{table}[t]
    \centering
    {
    \small
    \setlength{\tabcolsep}{3.6pt}
    \begin{tabular}{lcccc}
        \toprule
        Variant
        & Selection
        & BG update
        & Planner
        & SR (\(\uparrow\)) \\
        \midrule
        w/o DRS
        & Random
        & FDBU
        & EB-CEM
        & 76\% \\

        w/o FDBU
        & DRS
        & Copy
        & EB-CEM
        & 56\% \\

        w/o EB-CEM
        & DRS
        & FDBU
        & CEM
        & 82\% \\

        \textbf{Scope-WM}
        & DRS
        & FDBU
        & EB-CEM
        & \textbf{88\%} \\
        \bottomrule
    \end{tabular}
    }
    \caption{
        Component ablation on Push-T under the \(32\)-token full-dynamics
        setting used by Scope-WM (Faster).
    }
    \label{tab:component_ablation}
\end{table}

\begin{figure}[t]
    \centering
    \includegraphics[width=0.98\columnwidth]{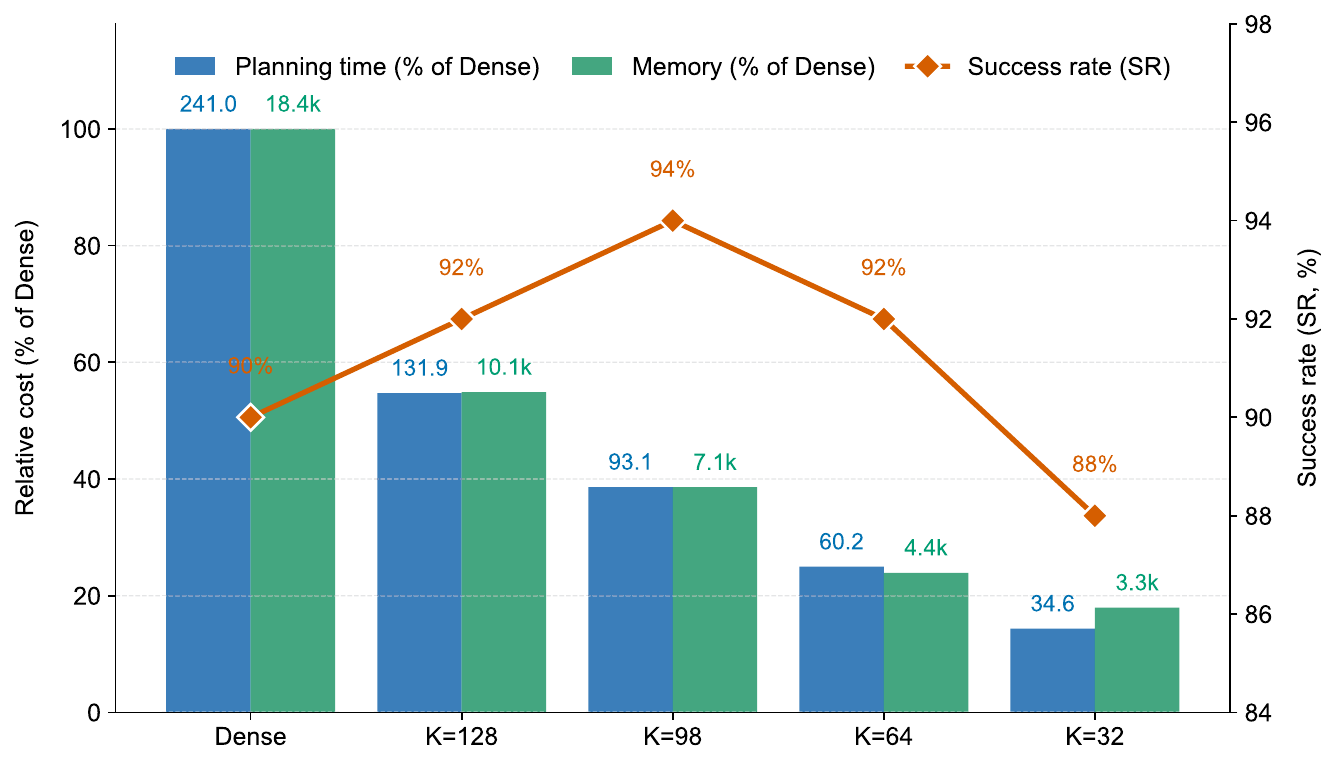}
    \caption{
        Effect of the full-dynamics token budget on Push-T.
        Bars show planning time and peak GPU memory normalized to dense
        DINO-WM, with absolute values annotated above the bars.
        The line shows success rate as the token budget varies across configurations.
    }
    \label{fig:token_scaling}
\end{figure}

\paragraph{Effect of token budget.}
We vary the full-dynamics token budget on Push-T to characterize the
performance--efficiency trade-off.
As shown in Figure~\ref{fig:token_scaling}, both planning time and peak
GPU memory decrease consistently as the token budget is reduced, while
SR remains between \(88\%\) and \(94\%\).
The \(K=98\) setting achieves the highest SR of \(94\%\) with only
\(38.6\%\) of the dense planning time and \(38.4\%\) of its peak
memory.
Even at \(K=32\), Scope-WM retains \(88\%\) SR while reducing these
costs to \(14.3\%\) and \(18.1\%\), respectively.

\begin{figure}[t]
    \centering
    \includegraphics[width=\columnwidth]{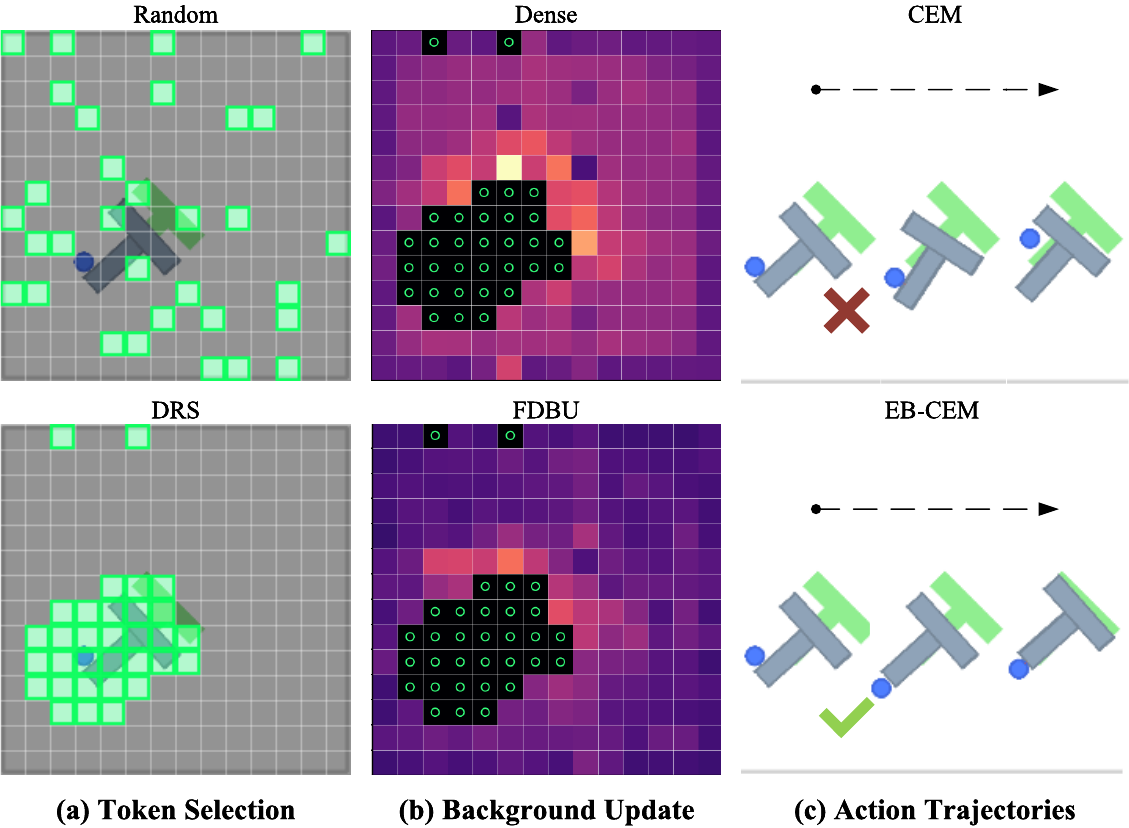}
    \caption{
        Qualitative analysis of Scope-WM on Push-T.
        (a) DRS focuses computation on prediction-relevant regions.
        (b) FDBU captures background evolution without dense
        token-to-token interactions.
        (c) EB-CEM reuses promising early trajectories for subsequent
        replanning.
    }
    \label{fig:qualitative_analysis}
\end{figure}

\begin{table}[t]
    \centering
    {
    \small
    \setlength{\tabcolsep}{10pt}
    \begin{tabular}{lcc}
        \toprule
        Selection
        & \multicolumn{2}{c}{Background update} \\
        \cmidrule(lr){2-3}
        &
        Copy
        & FDBU \\
        \midrule
        Random
        & 72\%
        & 76\% \\
        DRS
        & 56\%
        & 88\% \\
        \bottomrule
    \end{tabular}
    }
    \caption{
        Interaction between DRS and FDBU on Push-T under the same
        32-token budget and EB-CEM planner.
        Entries report success rate (SR, \(\uparrow\)).
    }
    \label{tab:drs_fdbu_interaction}
\end{table}

\paragraph{Interaction between DRS and FDBU.}
DRS determines which tokens receive full dynamics prediction, while
FDBU propagates foreground dynamics to the remaining latent state.
Table~\ref{tab:drs_fdbu_interaction} reveals an interaction
between the two mechanisms.
With random selection, FDBU provides a modest improvement from
72\% to 76\% SR.
In contrast, under DRS, replacing background copying with FDBU improves
SR substantially from 56\% to 88\%.
This suggests that concentrating full dynamics prediction on
prediction-relevant tokens makes effective background propagation
particularly important, revealing a strong complementarity between
DRS and FDBU.

\begin{table}[t]
    \centering
    {
    \small
    \setlength{\tabcolsep}{4.5pt}
    \begin{tabular}{llccc}
        \toprule
        Model
        & Planner
        & Samples
        & SR (\(\uparrow\))
        & Plan time (s, \(\downarrow\)) \\
        \midrule
        DINO-WM
        & CEM
        & 100/100
        & 90\%
        & 240.96 \\

        DINO-WM
        & EB-CEM
        & 300/100
        & 84\%
        & 267.36 \\
        \midrule
        Scope-WM
        & CEM
        & 100/100
        & 82\%
        & 30.67 \\

        Scope-WM
        & CEM
        & 200/200
        & 88\%
        & 60.96 \\

        Scope-WM
        & EB-CEM
        & 300/100
        & 88\%
        & 34.56 \\
        \bottomrule
    \end{tabular}
    }
    \caption{
        Interaction between sparse prediction and planning strategy on Push-T.
        Scope-WM uses the 32-token setting; sample budgets denote
        initial/subsequent MPC steps.
    }
    \label{tab:sparse_ebcem_interaction}
\end{table}

\begin{table}[t]
    \centering
    {
    \small
    \setlength{\tabcolsep}{4pt}
    \begin{tabular}{lcccc}
        \toprule
        & \multicolumn{3}{c}{CEM} & EB-CEM \\
        \cmidrule(lr){2-4}
        Early samples
        & 100 & 100 & 300 & 300 \\
        Later samples
        & 100 & 300 & 100 & 100 \\
        SR (\(\uparrow\))
        & 82\% & 82\% & 86\% & 88\% \\
        \bottomrule
    \end{tabular}
    }
    \caption{
        Effect of sample allocation and elite reuse on Push-T.
        Early and later denote initial and subsequent MPC steps.
    }
    \label{tab:ebcem_analysis}
\end{table}

\paragraph{Interaction between sparse prediction and EB-CEM.}
Sparse prediction and EB-CEM act on complementary sources of planning
cost: sparse prediction reduces the cost of each imagined rollout,
whereas EB-CEM reallocates search effort across MPC steps.
As shown in Table~\ref{tab:sparse_ebcem_interaction}, applying EB-CEM
to dense DINO-WM provides no benefit, decreasing SR from \(90\%\) to
\(84\%\) while increasing planning time from \(240.96\) to \(267.36\)~s.
For Scope-WM, standard CEM with 100 samples requires only \(30.67\)~s
but reaches \(82\%\) SR.
Increasing the budget to 200 samples recovers \(88\%\) SR at a cost of
\(60.96\)~s, whereas EB-CEM achieves the same \(88\%\) SR in only
\(34.56\)~s.
Thus, cheap sparse rollouts make front-loaded exploration affordable,
while elite reuse avoids maintaining a large sampling budget throughout
subsequent replanning.

\paragraph{Qualitative analysis.}
Figure~\ref{fig:qualitative_analysis} provides further intuition behind
the three designs under the aggressive 32-token setting.
Compared with random selection, DRS focuses full dynamics prediction on
prediction-relevant regions across the latent field (a).
FDBU achieves background prediction errors comparable to dense updating
using only the foreground dynamics context, suggesting that dense
token-to-token interactions are unnecessary for background propagation
(b).
Finally, EB-CEM retains promising early trajectories to guide
subsequent replanning (c).
Table~\ref{tab:ebcem_analysis} further confirms the benefits of front-loaded sampling and elite reuse.

\section{Conclusion}
\label{sec:conclusion}

We present Scope-WM, an efficient visual world-model planning framework that scopes computation across token selection, latent dynamics, and trajectory optimization.
Scope-WM combines DRS for prediction-aware token selection, FDBU for lightweight background updates, and EB-CEM for efficient trajectory optimization.
Together, these components reduce both the cost of individual imagined rollouts and the rollout budget required during subsequent replanning, while retaining a simple training pipeline consisting of one-off selector distillation followed by joint sparse world-model training.
Experiments demonstrate a favorable performance--efficiency trade-off while maintaining competitive planning performance.
On \textsc{Push-T}, the aggressive configuration achieves a \(6.97\times\) planning speedup while reducing planning time and peak GPU memory to \(14.3\%\) and \(18.1\%\) of dense DINO-WM, respectively, with only a modest change in task performance.
These results highlight computation allocation according to prediction and planning relevance as a promising principle for efficient visual world-model planning under flexible token and rollout budgets.

\bibliography{reference}

\clearpage
\appendix
\setcounter{secnumdepth}{2}

\setcounter{figure}{0}
\setcounter{table}{0}
\setcounter{equation}{0}
\renewcommand{\thefigure}{S\arabic{figure}}
\renewcommand{\thetable}{S\arabic{table}}
\renewcommand{\theequation}{S\arabic{equation}}

\twocolumn[{
    \centering
    {\LARGE\bfseries Supplementary Material\par}
    \vspace{1em}
}]

This technical supplement provides additional implementation,
evaluation, and experimental details for Scope-WM.
It follows the notation and task definitions of the main paper.

\section{Implementation and Reproducibility Details}
\label{sec:supp_implementation}

\subsection{Training Protocol and Checkpoint Selection}

All models are trained within a budget of at most \(100\) epochs.
Because convergence speed differs across tasks, the checkpoint used for
reporting is selected according to planning performance on the corresponding
validation split.
The Fast and Faster configurations are trained independently with
\(K=98\) and \(K=32\) full-dynamics tokens, respectively; Faster is not
obtained by changing the token budget only at planning time.

Training proceeds in two steps.
We first export gradient-times-input relevance targets from a trained dense
DINO-WM teacher and train a task-specific DRS head.
The distilled selector is then frozen, after which the sparse dynamics
predictor and FDBU are optimized jointly using the full-grid latent prediction
loss described in the main paper.
No separate training stage is used for FDBU.

\subsection{Model and Optimization Settings}

Table~\ref{tab:supp_model_config} summarizes the shared architecture and
optimization settings.
The frozen DINOv2 ViT-S/14 encoder produces \(196\) patch tokens from each
\(224\times224\) observation.
The visual decoder is disabled during model training and planning; it is used
only for the qualitative visualization reported below.

\begin{table*}[t]
    \centering
    {
    \small
    \setlength{\tabcolsep}{5.0pt}
    \begin{tabular}{lll}
        \toprule
        Component & Hyperparameter & Setting \\
        \midrule
        Visual representation
        & Encoder
        & Frozen DINOv2 ViT-S/14 \\

        & Input resolution
        & \(224\times224\) \\

        & Visual tokens / token dimension
        & \(196 / 384\) \\

        \midrule
        Dynamics predictor
        & Architecture
        & 6-layer causal Transformer \\

        & Attention heads / head dimension
        & \(16 / 64\) \\

        & Feed-forward hidden dimension
        & \(2048\) \\

        & Transformer / embedding dropout
        & \(0.1 / 0\) \\

        & Action / proprioceptive embedding dimension
        & \(10 / 10\) \\

        & Action/proprio standardization
        & Dataset statistics \\

        & Prediction steps per world-model call
        & \(1\) \\

        \midrule
        Sparse world-model training
        & Optimizer
        & AdamW \\

        & Predictor and action-encoder learning rate
        & \(1\times10^{-4}\) \\

        & Weight decay
        & \(0.01\) \\

        & Maximum training epochs
        & \(100\) \\

        & Latent supervision
        & MSE over all \(196\) latent positions \\

        & Training seed
        & \(0\) \\

        \midrule
        FDBU
        & Foreground summaries
        & Mean predicted state and mean foreground delta \\

        & Context projection
        & LN--Linear--GELU \\

        & Update branch
        & LN--Linear--GELU--Linear \\

        & Gate branch
        & LN--Linear--Sigmoid \\

        & Hidden multiplier
        & \(2.0\) \\

        & Residual scale \(\alpha\)
        & \(1.0\) \\

        & Dropout
        & \(0\) \\
        \bottomrule
    \end{tabular}
    }
    \caption{
        Shared model and optimization settings.
        LN denotes layer normalization.
    }
    \label{tab:supp_model_config}
\end{table*}

\subsection{DRS Distillation}

For each of the five tasks, DRS is distilled from the corresponding dense
DINO-WM teacher.
The teacher relevance target is computed from the full-grid latent prediction
loss using the gradient-times-input score defined in the main paper.
The selector receives each globally contextualized DINOv2 patch token together
with the action and proprioceptive embeddings.
We use the same target-export procedure, selector architecture, and
distillation configuration across both rigid and deformable tasks, as
summarized in Table~\ref{tab:supp_drs_config}.
With a batch size of \(8\) and at most \(400\) exported batches, each task
uses up to \(3{,}200\) trajectory windows for selector distillation.

\begin{center}
    {
    \small
    \setlength{\tabcolsep}{5.0pt}
    \begin{tabular}{lc}
        \toprule
        Hyperparameter & Setting \\
        \midrule
        Target type & Gradient-times-input \\
        Target split & Training split \\
        Export batch size & \(8\) \\
        Maximum exported batches & \(400\) \\
        Target-export workers & \(8\) \\
        Selector hidden dimension & \(128\) \\
        Distillation epochs & \(80\) \\
        Distillation batch size & \(256\) \\
        Optimizer & AdamW \\
        Learning rate & \(3\times10^{-4}\) \\
        Weight decay & \(1\times10^{-4}\) \\
        Objective & KL divergence \\
        Temperature \(T\) & \(1.0\) \\
        Action/proprio conditioning & Enabled \\
        Selector during sparse-WM training & Frozen \\
        \bottomrule
    \end{tabular}
    }
    \captionof{table}{
        DRS target-export and distillation settings shared across all five tasks.
    }
    \label{tab:supp_drs_config}
\end{center}

\subsection{Task-Specific Training Settings}

Table~\ref{tab:supp_task_training} reports the task-specific settings.
PointMaze and Push-T use three observation-history frames and a
frameskip of five.
Wall uses one observation-history frame with the same frameskip of five.
Rope and Granular use one history frame and no action subsampling.
All available trajectories in the configured training directories are used.
PointMaze, Wall, Rope, and Granular use a \(90/10\) trajectory split generated
with seed \(42\), while Push-T uses the predefined training and validation
directories distributed with the dataset.

\begin{table*}[t]
    \centering
    {
    \small
    \setlength{\tabcolsep}{3.9pt}
    \begin{tabular}{lcccccccc}
        \toprule
        Task
        & History
        & Frameskip
        & Global batch
        & Training GPUs
        & Epoch budget
        & Train \(K\)
        & Eval. instances
        & Max MPC steps \\
        \midrule
        PointMaze
        & 3 & 5 & 512 & 8 V100
        & \(\leq100\) & 98 / 32 & 50 & 15 \\

        Wall
        & 1 & 5 & 512 & 8 V100
        & \(\leq100\) & 98 / 32 & 50 & 15 \\

        Push-T
        & 3 & 5 & 512 & 8 V100
        & \(\leq100\) & 98 / 32 & 50 & 15 \\

        Rope
        & 1 & 1 & 64 & 1 RTX 2080 Ti
        & 100 & 98 / 32 & 10 & 7 \\

        Granular
        & 1 & 1 & 64 & 1 RTX 2080 Ti
        & 100 & 98 / 32 & 10 & 7 \\
        \bottomrule
    \end{tabular}
    }
    \caption{
        Task-specific training and evaluation settings.
        Fast and Faster are independently trained with \(K=98\) and \(K=32\),
        respectively.
    }
    \label{tab:supp_task_training}
\end{table*}

\subsection{Planning Configuration}

Unless otherwise stated, standard CEM uses \(100\) candidates, \(10\) elites,
and \(10\) optimization iterations.
EB-CEM uses \(300\) candidates and \(30\) elites at the initial MPC step,
followed by \(100\) candidates and \(10\) elites at subsequent steps.
A bank of \(30\) candidates is saved after the initial MPC search.
At later steps, \(70\%\) of the population is sampled locally around banked
elites with perturbation scale \(0.5\), while the remaining \(30\%\) is drawn
from the current CEM Gaussian.
The elite bank is initialized once and is not shifted or replaced during later
MPC steps.
The visual decoder is not loaded during planning.

\begin{center}
    {
    \small
    \setlength{\tabcolsep}{4.7pt}
    \begin{tabular}{lc}
        \toprule
        Hyperparameter & Setting \\
        \midrule
        Planning horizon & 5 \\
        CEM optimization iterations & 10 \\
        Initial variance scale & 1.0 \\
        Standard population / elites & \(100 / 10\) \\
        Initial EB-CEM population / elites & \(300 / 30\) \\
        Later EB-CEM population / elites & \(100 / 10\) \\
        Elite-bank size \(M_b\) & 30 \\
        Bank-save MPC step & 0 \\
        Bank-use start step & 1 \\
        Local fraction \(\eta\) & 0.7 \\
        Global fraction & 0.3 \\
        Perturbation scale \(\lambda\) & 0.5 \\
        Sequence shift & 0 \\
        Bank update mode & Initial only \\
        Planning objective & Final predicted latent \\
        \bottomrule
    \end{tabular}
    }
    \captionof{table}{
        CEM and EB-CEM hyperparameters used for the main Scope-WM results.
    }
    \label{tab:supp_planning_config}
\end{center}

The planning objective is evaluated on the final predicted latent state.
PointMaze uses the visual-latent cost only (objective weight \(\alpha=0\)),
whereas Wall, Push-T, Rope, and Granular use equal visual and
proprioceptive weights (\(\alpha=1\)).
PointMaze, Wall, Rope, and Granular construct goals from random environment
states, while Push-T uses a five-step segment from the validation data.

\FloatBarrier

\section{Evaluation Protocol}
\label{sec:supp_evaluation}

\subsection{Datasets and Goal Construction}

We follow the task definitions and offline datasets used by DINO-WM.
All observations are resized and center-cropped to \(224\times224\), then
normalized channel-wise with mean and standard deviation \(0.5\).
Actions and proprioceptive states are standardized using dataset statistics.
PointMaze and Wall use procedurally sampled initial--goal pairs under fixed
evaluation seeds.
Push-T draws a segment of length \(5\times5+1\) simulator frames from the
validation set: the first frame is used as the initial state, and the recorded
low-level actions are replayed to construct the five-step goal observation.

For Rope and Granular, the initial particle state is drawn from the validation
data.
The goal is generated using the environment's task-specific geometric
transformation.
For Rope, the goal applies a random in-plane rotation and translation.
For Granular, the goal applies a random scale and translation.
Planning compares frozen-encoder latent features, whereas the final evaluation
uses the simulator state.

\subsection{Random Seeds and Number of Runs}

Sparse world-model training uses seed \(0\), and random trajectory splits use
seed \(42\).
Planning follows the official DINO-WM seed-generation rule.
With planning seed \(99\), evaluation instance \(i\) uses
\[
    s_i = 99i + 1,\qquad i=0,\ldots,n-1.
\]
PointMaze, Wall, and Push-T are evaluated on \(50\) fixed instances, while Rope
and Granular are evaluated on \(10\) fixed instances.
Every reported Scope-WM result is obtained from one selected checkpoint
evaluated on the complete fixed instance set; results are not averaged across
independently trained models.
Within each task, all locally evaluated variants use the same evaluation seeds
and initial--goal construction procedure.

\subsection{Metrics and Efficiency Measurement}

PointMaze and Wall count an episode as successful when the final positional
distance to the goal is below \(0.5\) and \(4.5\), respectively.
Push-T requires both a state-position distance below \(20\) and an orientation
error below \(\pi/9\).
For Rope and Granular, we report the symmetric Chamfer Distance between the
final and goal particle sets using their three-dimensional coordinates:
\[
\begin{aligned}
\operatorname{CD}(X,Y)
={}&\frac{1}{|X|}\sum_{x\in X}\min_{y\in Y}\lVert x-y\rVert_2\\
&+\frac{1}{|Y|}\sum_{y\in Y}\min_{x\in X}\lVert y-x\rVert_2.
\end{aligned}
\]

Planning time is measured as the cumulative CEM sub-planner time over all
executed MPC steps.
Environment stepping, metric computation, decoding, and visualization are
excluded.
Peak GPU memory is the maximum CUDA memory allocated during a complete
evaluation run after resetting the peak-memory counter.
FLOPs are measured for one world-model prediction and reported per prediction
sample.

The cross-task task-performance values of prior methods in the main paper are
taken from their published reports.
For the local Push-T efficiency comparison, dense DINO-WM and Scope-WM are
evaluated on the same hardware and evaluation instances.
The dense local baseline uses standard CEM with \(100\) candidates,
\(10\) elites, and \(10\) optimization iterations.

\section{Hardware and Software Environment}
\label{sec:supp_hardware}

\paragraph{PointMaze, Wall, and Push-T.}
These models are trained using eight NVIDIA V100 GPUs, while planning
evaluation is performed on a single 32-GB NVIDIA V100 GPU.
The single-GPU platform provides 16 virtual CPU cores from an Intel Xeon
Platinum 8352V processor at 2.10\,GHz and 64\,GB of host memory, and runs
Ubuntu 22.04.
The primary software environment uses Python 3.9.19, PyTorch 2.3.0,
torchvision 0.18.0, CUDA 12.1, and cuDNN 8.9.2.
We use Accelerate 0.26.1 for distributed training, Hydra 1.2.0 and
OmegaConf 2.3.0 for configuration management, and NumPy 1.26.4.
The corresponding simulation dependencies include Gym 0.23.1, D4RL 1.1,
MuJoCo 3.2.7, and mujoco-py 2.1.2.14.

\paragraph{Rope and Granular.}
The deformable tasks are trained and evaluated on one NVIDIA GeForce RTX
2080 Ti GPU with 22,528\,MiB of memory.
The machine provides an Intel Xeon Platinum 8255C CPU at 2.50\,GHz and
503\,GiB of host memory, and runs Ubuntu 22.04.3 LTS.
The Python environment uses Python 3.12.3 and PyTorch 2.3.0+cu121 with
CUDA 12.1 and cuDNN 8.9.2.
The PyFlex simulator bindings are compiled separately with the CUDA 9.2
toolkit required by the available source build and loaded into the PyTorch
environment through the paths provided in the code supplement.

\section{Additional Efficiency Results}
\label{sec:supp_efficiency}

\subsection{Efficiency Across Rigid-Body Tasks}

Table~\ref{tab:supp_cross_task_efficiency} extends the end-to-end efficiency
comparison from Push-T to PointMaze and Wall.
For each task, dense DINO-WM and Scope-WM are measured on the same GPU
using the same initial--goal pair and a forced full-length planning run.
Planning time denotes the cumulative CEM sub-planner time across all MPC
steps, consistent with the metric used in the main paper.
Task success rates are taken from the corresponding complete evaluation
sets reported in the main paper.

\begin{table*}[t]
    \centering
    {
    \small
    \setlength{\tabcolsep}{5.0pt}
    \begin{tabular}{llcccccc}
        \toprule
        Task
        & Method
        & \(K\)
        & Plan time (s, \(\downarrow\))
        & Speedup (\(\uparrow\))
        & Peak Mem. (MB, \(\downarrow\))
        & Mem. Ratio (\(\downarrow\))
        & SR (\(\uparrow\)) \\
        \midrule

        PointMaze
        & DINO-WM
        & 196
        & 204.52
        & \(1.00\times\)
        & 6309
        & 100.0\%
        & 98\% \\

        PointMaze
        & Scope-WM (Faster)
        & 32
        & \textbf{35.01}
        & \(\mathbf{5.84}\times\)
        & \textbf{3334}
        & \textbf{52.8\%}
        & 98\% \\

        \midrule

        Wall
        & DINO-WM
        & 196
        & 59.44
        & \(1.00\times\)
        & \textbf{1542}
        & \textbf{100.0\%}
        & 96\% \\

        Wall
        & Scope-WM (Faster)
        & 32
        & \textbf{19.60}
        & \(\mathbf{3.03}\times\)
        & 1939
        & 125.8\%
        & 92\% \\

        \midrule

        Push-T
        & DINO-WM
        & 196
        & 240.96
        & \(1.00\times\)
        & 18436
        & 100.0\%
        & 90\% \\

        Push-T
        & Scope-WM (Faster)
        & 32
        & \textbf{34.56}
        & \(\mathbf{6.97}\times\)
        & \textbf{3334}
        & \textbf{18.1\%}
        & 88\% \\

        \bottomrule
    \end{tabular}
    }
    \caption{
        End-to-end efficiency on three rigid-body planning tasks.
        Time and peak memory use one fixed full-length run; SR uses the complete
        \(50\)-instance set.
        Memory ratios are normalized per task, and the peak includes the larger
        initial EB-CEM population.
    }
    \label{tab:supp_cross_task_efficiency}
\end{table*}

Scope-WM consistently reduces planning time across the three tasks,
achieving \(5.84\times\), \(3.03\times\), and \(6.97\times\) speedups on
PointMaze, Wall, and Push-T, respectively.
The peak-memory benefit depends on the relative contributions of rollout
activations and search-population overhead.
Scope-WM substantially reduces memory on PointMaze and Push-T, where dense
latent rollouts incur larger activation footprints.
On Wall, the dense baseline already has a small memory footprint, while
EB-CEM temporarily uses a larger population at the initial MPC step.
Consequently, its initial-search overhead outweighs the smaller savings from
token sparsity, producing a modest absolute memory increase.
These results indicate that scoped latent prediction provides the largest
memory savings when dense rollout activations dominate the overall footprint,
while its planning-time benefit remains consistent across tasks.

\section{Additional Qualitative Results}
\label{sec:supp_qualitative}

\begin{figure*}[t]
    \centering
    \includegraphics[width=\textwidth]{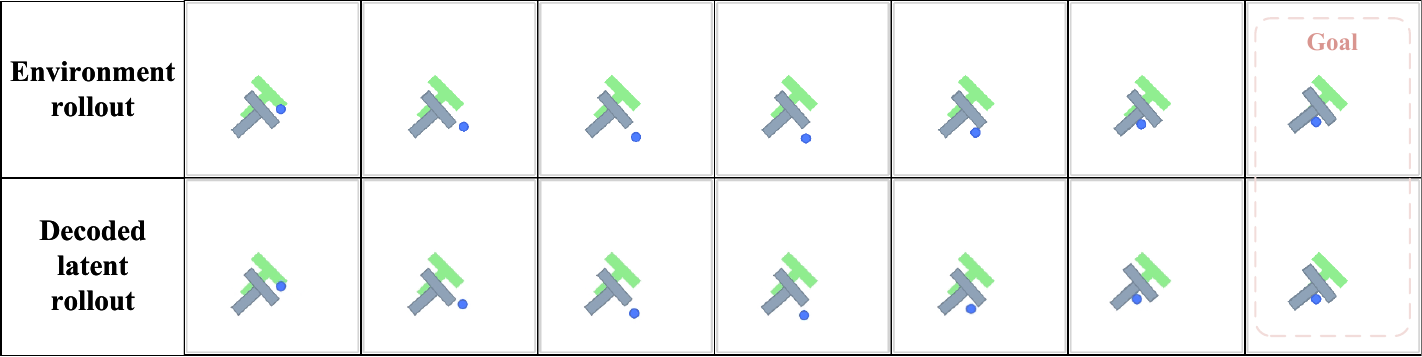}
    \caption{
        Environment and decoded latent rollouts on Push-T.
        The upper row executes the planned actions in the environment; the lower
        row decodes the corresponding Scope-WM latent rollout with the dense
        DINO-WM decoder.
        Steps proceed left to right, and the highlighted final column is the goal.
        The decoder is used only for visualization.
    }
    \label{fig:supp_decoded_rollouts}
\end{figure*}

\paragraph{Decoded latent rollouts.}
Figure~\ref{fig:supp_decoded_rollouts} compares a Scope-WM latent rollout
with the corresponding environment rollout under the same planned action
sequence.
The decoded rollout follows the overall motion of the agent and the
T-shaped object across successive steps, indicating that the sparse latent
dynamics preserve task-relevant motion information.
The decoder is used only for visualization and is not loaded during planning.

\FloatBarrier

\section{Additional Token-Budget Analysis}
\label{sec:supp_token_budget}

Fast and Faster are trained independently with \(K=98\) and \(K=32\),
respectively.
To examine whether the full-dynamics token budget can instead be changed only
at planning time, we evaluate each checkpoint under both token budgets.
As shown in Table~\ref{tab:supp_topk_mismatch}, changing \(K\) after training
substantially degrades planning performance in both directions.

\begin{center}
    
    {
    \small
    \setlength{\tabcolsep}{8.0pt}
    \begin{tabular}{ccc}
        \toprule
        Training \(K\)
        & Planning \(K\)
        & SR (\(\uparrow\)) \\
        \midrule
        98 & 98 & \textbf{94\%} \\
        98 & 32 & 14\% \\
        \midrule
        32 & 32 & \textbf{88\%} \\
        32 & 98 & 54\% \\
        \bottomrule
    \end{tabular}
    }
    \captionof{table}{
        Training--planning token-budget consistency on Push-T.
        Each configuration uses the same \(50\) fixed instances.
    }
    \label{tab:supp_topk_mismatch}
\end{center}

The degradation indicates that \(K\) is part of the learned sparse-dynamics
configuration rather than a freely adjustable inference-time parameter.
The sparse predictor and FDBU are jointly optimized under a particular
foreground--background partition: changing \(K\) at planning time alters both
the number of tokens receiving full dynamics prediction and the distribution
of tokens handled by the background update.
This creates a mismatch from the computation pattern observed during
training.
Accordingly, we train separate Fast and Faster models for their respective
planning budgets.

\end{document}